\documentclass[letterpaper]{article} % DO NOT CHANGE THIS
\usepackage{aaai2026}  % DO NOT CHANGE THIS
\usepackage{times}  % DO NOT CHANGE THIS
\usepackage{helvet}  % DO NOT CHANGE THIS
\usepackage{courier}  % DO NOT CHANGE THIS
\usepackage{amssymb}
\usepackage[hyphens]{url}  % DO NOT CHANGE THIS
\usepackage{graphicx} % DO NOT CHANGE THIS
\usepackage{natbib}  % DO NOT CHANGE THIS AND DO NOT ADD ANY OPTIONS TO IT
\usepackage{caption} % DO NOT CHANGE THIS AND DO NOT ADD ANY OPTIONS TO IT
\usepackage{algorithm}
\usepackage{algorithmic}

\usepackage{newfloat}
\usepackage{listings}

\usepackage{marvosym}

\DeclareCaptionStyle{ruled}{labelfont=normalfont,labelsep=colon,strut=off} % DO NOT CHANGE THIS
\floatstyle{ruled}
\newfloat{listing}{tb}{lst}{}
\floatname{listing}{Listing}

\usepackage{amsmath}
\usepackage{amssymb}
\usepackage{amsthm}
\usepackage{booktabs}
\usepackage{enumitem}
\usepackage{tikz}
\usepackage{pgfplots}
\pgfplotsset{compat=1.18}

\title{Next Token Knowledge Tracing: Exploiting Pretrained LLM Representations to Decode Student Behaviour.}

\author{
    Max Norris$^1$, Kobi Gal$^1$ and Sahan Bulathwela$^2$
}

\affiliations{
    $^1$ University of Edinburgh, The United Kingdom \\
    $^2$ Centre for Artificial Intelligence, University College London, The United Kingdom
}

\begin{document}

\maketitle

\begin{abstract}
Modelling student knowledge is a key challenge when leveraging AI in education, with major implications for personalised learning. The Knowledge Tracing (KT) task aims to predict how students will respond to educational questions in learning environments, based on their prior interactions. Existing KT models typically use response correctness along with metadata like skill tags and timestamps, often overlooking the question text—an important source of pedagogical insight. This omission poses a lost opportunity while limiting predictive performance. We propose \emph{Next Token Knowledge Tracing (NTKT)}, a novel approach that reframes KT as a next-token prediction task using pretrained Large Language Models (LLMs). NTKT represents both student histories and question content as sequences of text, allowing LLMs to learn patterns in both behaviour and language. Our series of experiments significantly improves performance over state-of-the-art neural KT models and generalises much better to cold-start questions and users. These findings highlight the importance of question content in KT and demonstrate the benefits of leveraging pretrained representations of LLMs to model student learning more effectively.
\end{abstract}

\section{Introduction}

Modelling latent variables such as student knowledge \cite{truelearn, qiu2024toolbox}, skills and learning progression is a core challenge in the design of %Intelligent Tutoring Systems
Student Models \cite{mousavinasab2021intelligent}, that make up a key part of intelligent learning companions of the future \cite{perez2021ai}. Knowledge Tracing (KT) addresses this by using students' responses to questions, as observable signals of their underlying knowledge state. A wide range of KT models have been developed to predict student performance on future questions \cite{khajah2014integrating, cen2006learning, pavlik2009performance, pandey2019self}. More recent approaches leverage deep learning to model complex patterns in student interaction histories, enabling more accurate response prediction \cite{piech2015deep, liu2019ekt}.

Deep learning approaches to Knowledge Tracing (KT) have demonstrated considerable potential, yet their progress is limited by dataset constraints. Many widely used open-source KT datasets (e.g., ASSISTments, EdNet \cite{feng2009addressing,choi2020ednet}) exclude the text of the questions answered by students \cite{choi2020ednet}, a common omission in large datasets released by private companies due to intellectual property restrictions.

As a result, most existing KT models have been built around architectures that do not incorporate natural language features. These models typically rely on discrete categorical inputs to represent questions, such as subject, difficulty level, and concept or skill tags. This paradigm, known as the ``ID-based approach" \cite{yuan2023go}, requires extensive manual annotation by domain experts to assign these features to each question. While often absent in public datasets, question text is always available to stakeholders aiming to deploy personalisation technologies internally within their learning platforms. Hence, exploiting the feasibility of exploiting linguistic representations is an obvious direction demanding exploration.

The text of a question often carries rich pedagogical signals that could enhance predictive accuracy in Knowledge Tracing models. For example, Barbu et al. (2010) found that linguistic complexity affects student comprehension in math problem-solving, showing how textual features influence performance in real world settings \cite{barbu2010effects}. More broadly, full question text carries finer-grained information about topics, structure, and wording that simple categorical labels fail to capture. Reducing the question to these coarser-grain annotations discards this nuance, limiting the model's ability to access useful signals or even learn new ones. Natural language is also helpful when learning generalizable representations in the question space; this has benefits for cold-start scenarios where students encounter unseen questions, a challenge for current KT systems \cite{jung2024clst}. 

We hypothesise that effectively incorporating question text alongside interaction history metadata can lead to improved knowledge tracing performance and better cold-start generalisation. To this end, we propose a novel approach that tightly integrates question text into the KT objective: what we term Next Token Knowledge Tracing (NTKT). Leveraging the strengths of LLMs in natural language understanding \cite{jin2023large,singhal2023large,zhao2023recommender}, NTKT reformulates knowledge tracing as a next-token prediction task. Student interaction histories are converted into natural language sequences, allowing fine-tuned LLMs to predict the next token i.e., the student's response—based on their full contextual history, including question text.

To validate our approach, we conduct three experiments. First, we benchmark NTKT against existing Deep Learning baselines, demonstrating improved predictive accuracy. Second, we conduct controlled ablation studies to isolate the impact of textual information by comparing model performance across varying input representations, from full question text to ID-only formats. Finally, we assess NTKT's generalisation capabilities under two cold-start scenarios, 1) user cold start and 2) question cold start. 

This work makes three contributions. First, it introduces a new framework that reformulates knowledge tracing as a next-token prediction task, enabling the direct use of question content and overcoming limitations of ID-based models. Second, it offers insights into which textual elements influence predictions, contributing to model transparency through feature importance. Third, it demonstrates improved generalisation in cold-start settings, a persistent challenge for conventional KT approaches. These contributions lay the groundwork for research in text-aware educational AI \footnote{Link to prompts and code published upon acceptance.}. 
% We will open-source all prompts, fine-tuning code and models to the community.

\section{Related Work}

Our work lies at the intersection of two key research areas: Knowledge Tracing and Natural Language Processing.
% In the following section we outline relevant literature from both fields respectively.

\subsection{Knowledge Tracing}

Knowledge Tracing involves modelling a student’s evolving knowledge over time, using their interaction history to predict future performance. This predictive ability enables adaptive learning systems to tailor instruction and feedback.

Early KT approaches used statistical models such as Bayesian Knowledge Tracing (BKT) \cite{khajah2014integrating,pardos2010modeling,yudelson2013individualized}, which represent knowledge states as binary variables and offer interpretability and simplicity. Other models like Learning Factor Analysis (LFA) and Performance Factor Analysis (PFA) \cite{cen2006learning,pavlik2009performance} extend this approach using logistic regression to model student performance and learning.

Deep learning has since reshaped KT. Deep Knowledge Tracing (DKT) \cite{piech2015deep} introduced RNNs for sequence modelling. Following this work was a range of extensions starting with models such as DKVMN \cite{zhang2017dynamic}, which incorporated memory networks. Next, a range of attention-based models, including SAKT \cite{pandey2019self}, AKT \cite{ghosh2020context} and RKT \cite{pandey2020rkt}, demonstrated improved performance on long sequences. Other recent state-of-the-art methods include the DTransformer \cite{yin2023tracing}, which uses a contrastive loss for stable knowledge tracing.

Several studies have explored the incorporation of NLP in KT to make use of question text. Early approaches like EKT \cite{liu2019ekt} leveraged models such as word2vec \cite{church2017word2vec} and Bert \cite{devlin2018bert} to generate pretrained embeddings of question text that were later incorporated alongside existing KT architectures.

 More recently, LLMs have been applied to KT via prompting \cite{neshaei2024towards}, multi-agent setups \cite{trifa2019knowledge}, and fine-tuning on partial student histories \cite{jung2024clst}. These approaches undermine the value of domain-adaptation of content representations, whereas our work shows the performance gain through finetuning of LLMs to KT. 
 
The language model–based knowledge tracing method (LKT) has been shown to apply smaller encoder-based pre-trained models effectively to KT in a masked language modelling setup \cite{lee2024language}. While this method boasts impressive performance, it is limited by short context windows (512 tokens), capturing only a few student interactions at a time—despite students often having long interaction histories. Furthermore, their experiments are based on datasets not widely used in the KT community. Crucially, this method relies on smaller encoder-based models with a masked language modelling objective, in contrast to the more recent advancements with much larger decoder-only architectures.

\subsection{NLP Tasks}

Recent advances in NLP have been driven by decoder-only large language models (LLMs), which generate text by predicting the next token in a sequence. These models, such as GPT-3 \cite{brown2020language} and LLaMA \cite{touvron2023llama}, have achieved strong performance across a wide range of tasks, demonstrating that next-token prediction is a powerful pretraining objective. Furthermore, fine-tuned LLMs have been shown to outperform traditional approaches on many classification and reasoning benchmarks \cite{luo2025large,sui2503stop}, suggesting that their pretrained representations can be adapted for structured tasks like Knowledge Tracing.

Accordingly, this research examines the potential of fine-tuning pretrained decoder only LLMs for KT. Specifically, we focus on open-source models such as the LLaMA family due to their strong performance, open-source nature, and compatibility for finetuning. 
% ASR literature has set a precedent for loss masking. we borrow
However, adapting generative LLMs to classification tasks can introduce noise, as all output tokens typically contribute to the loss. To address this, we adopt a selective loss masking strategy, computing loss only at target positions (e.g., predicted correctness tokens). This simplifies optimization by aligning the supervision signal with the task. 

Similar masking strategies have been used to denoise training in ASR \cite{chen2023meditron} and sequence labeling tasks \cite{dukic2024looking}, showing improved learning efficiency and alignment.

\section{Methodology}

In this section, we describe how KT can be formulated as a next token prediction problem. We first formalise the problem setup and notation, followed by our proposed model architecture and training procedure.

\subsection{Research Questions}

We structure our investigation to systematically answer a series of research questions, leading to evaluating the value of exploiting pretrained representations for the KT task. The research questions are listed below:

\begin{itemize}
    \item \textbf{RQ1:} Does finetuning LLMs on the KT task significantly improve performance against existing baselines?  
    % \item \textbf{RQ2:} Can fine-tuning exclusively to predict the answer outcomes improve computational efficiency without harming performance on KT?
    \item \textbf{RQ2:} Which exercise attribute is most important to improve predictive performance?
    \item \textbf{RQ3:} How do pretrained representations contribute to addressing the cold-start problem?
    % \item \textbf{RQ5:} 
\end{itemize}

\subsection{Problem Setup}

Each learner $\ell \in \{1, \dots, L\}$ interacts with a sequence of exercises over 
discrete timesteps $t = 1, \dots, T_\ell$. At timestep $t$, learner $\ell$ attempts a multiple-choice exercise $e_t^\ell = (q_t^\ell, O_t^\ell)$,
where $q_t^\ell$ denotes the question and 
$O_t^\ell = \{ o_{t_1}^\ell, \dots, o_{t_J}^\ell \}$ is the set of answer options.
The learner's binary correctness outcome is denoted $a_t^\ell \in \{0,1\}$.
We denote the learner's interaction history up to time $t$ as
\begin{equation}
\label{eq:history}
\mathcal{H}_t^\ell
\;=\;
\bigl\{ (e_1^\ell, a_1^\ell), \dots, (e_{t-1}^\ell, a_{t-1}^\ell) \bigr\}.
\end{equation}
% but in equation block
Following the standard knowledge tracing formulation, the objective at 
timestep $t$ is to predict whether learner $\ell$ will answer the current exercise 
correctly. Given the current exercise $e_t^\ell = (q_t^\ell, O_t^\ell)$ and the 
historical sequence $\mathcal{H}_t^\ell$, the predictive model estimates
\begin{equation}
\label{eq:response-prob}
P_\Theta\!\left(
a_t^\ell = 1
\;\middle|\;
q_t^\ell, O_t^\ell, \mathcal{H}_t^\ell
\right).
\end{equation}
where $P_\Theta$ denotes the model parameterized by $\Theta$.

\subsection{Dataset}

In this research we use the Eedi dataset, a large collection of multiple choice mathematics word problems from the NIPS 2020 Education Challenge~\cite{wang2020instructions}. 
It contains 1,905,115 interactions from 39,873 students across 1,657 unique exercises covering 544 distinct constructs/Knowledge Components. 
Student sequences have a maximum length of 2,712, with a mean of 47.8 and a median of 32. The overall correctness rate is 62.3\%.

These exercises typically present mathematical problems in natural language. An example of an exercise in the data set is:

\begin{center}
    \textit{Asha's office is on level 7. Her car is parked in the basement on level -2. How many floors does Asha need to go down to get from the office to her car?}
\end{center}

\begin{center}
    \textbf{A) 10} \hspace{0.5cm} \textbf{B) 5} \hspace{0.5cm} \textbf{C) 9} \hspace{0.5cm} \textbf{D) 7}
\end{center}

\subsection{Baseline Models}

To assess the performance of NTKT, we compare it against several established knowledge tracing architectures. These baselines represent the evolution of the field from recurrent neural networks to modern attention-based systems. We specifically adapted certain models to incorporate textual data to ensure a fair comparison with our text-aware approach.

\begin{description} 
    \item[\textbf{Deep Knowledge Tracing (DKT)}]\cite{piech2015deep} DKT utilizes Long Short-Term Memory (LSTM) networks to process a student's interaction history as a chronological sequence. The model maintains a compressed numerical representation, or hidden state, of a student's latent knowledge, which is updated at each timestep based on the specific question identifier and the student's previous correctness. The final output is a probability indicating the likelihood of the student answering the next question correctly.

    \item[\textbf{Attentive Knowledge Tracing (AKT)}]\cite{ghosh2020context} shifts away from recurrent architectures in favor of a self-attention mechanism. This model uses a Transformer-style approach to calculate the relevance between a student's past interactions and the current target question. Unlike recurrent models that compress history into a single cumulative state, AKT can directly "attend" to specific past experiences. It incorporates a temporal decay function to prioritize recent learning while using specialized embeddings to capture the relationship between different educational concepts.

    \item[\textbf{AKT-text}] is our augmented version of the AKT model, designed to evaluate if traditional KT architectures can be effectively retrofitted with access to full question text. We use a pre-trained sentence transformer (all-MiniLM-L6-v2) to convert the question text into a dense 384-dimensional numerical vector. This embedding is then projected and concatenated with the standard question and concept identifiers. This allows the AKT attention mechanism to factor in the semantic meaning of the question text alongside categorical metadata.

    \item[\textbf{DTransformer}]\cite{yin2023tracing} is a state-of-the-art Transformer-based architecture that introduces more stable knowledge tracking through a decoupled representation of questions and student proficiency. It employs a contrastive learning objective to ensure that similar learning states are mapped closely together in the model's internal representation space.
    \item[\textbf{No-FT (Zero-Shot LLaMA)}] serves as a control without domain-specific adaptation. We use the LLaMA 3.2 3B model in its base instructional state, providing it with the same history and target question as our NTKT approach but without any fine-tuning on student interaction data. This allows us to isolate the performance gains attributable specifically to our fine-tuning framework.
\end{description}

\subsection{NTKT Models: A Causal Language Modelling Approach to Knowledge Tracing}

\begin{figure}[t]
\centering
\includegraphics[width=\linewidth]{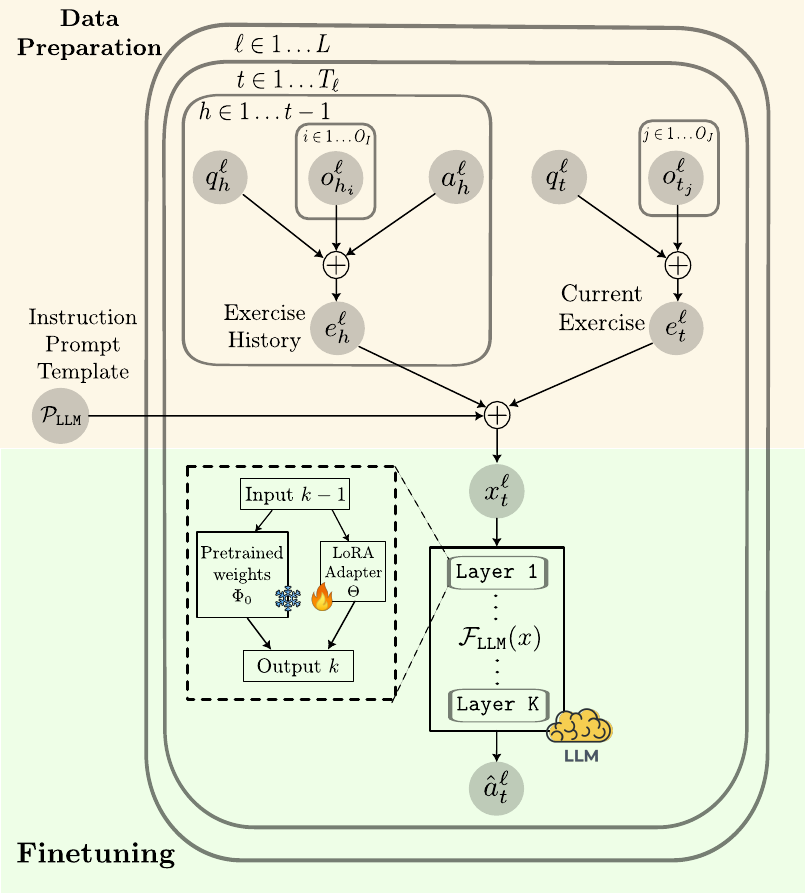}
\caption{NTKT Pipeline: Data Preparation (Orange) and Fine-tuning (Green) with Grey circles representing different observable variables in the dataset including the target variable. $\oplus$ represents the concatenation operator where variable values are concatenated together to create training examples. The LoRA training involves keeping LLM weights $\Phi_0$ frozen and training the adapter weights $\Theta$ for each layer. 
}
\label{fig:pipeline}
\end{figure}

NTKT reformulates knowledge tracing as a language modelling task, where student performance prediction is set up as a causal language modelling objective. We outline two phases of the modelling pipeline as detailed in Figure~\ref{fig:pipeline}.

\paragraph{Data Preparation}  
In the data preparation stage, we construct each input sequence \( x^{\ell}_t \) by combining a student's interaction history \( \mathcal{H}^{\ell}_{t} \), the target exercise \( e^{\ell}_{t} \), and an instruction prompt template \( \mathcal{P}_\texttt{LLM} \) (see Listing~1). 

\vspace{1cm}

\lstset{
  basicstyle=\ttfamily\small,
  breaklines=true,
  frame=single,
  columns=fullflexible
} \label{listing:prompt-template}

To ensure that the model can reliably distinguish between different components of the sequence, we annotate interaction records using XML-style tags—for example, \texttt{<Q>} \texttt{</Q>} to delimit questions previously answered by the student, and \texttt{<cr>} \texttt{</cr>} to indicate the correctness label (``Correct'' or ``Incorrect''). This tagging scheme follows established practices in both prompt engineering and fine-tuning workflows~\cite{dabre2024effective,anthropic_xml_tags}, and facilitates explicit structural cues for the language model. 

To refer to a specific token index within $x^{\ell}_{t}$, we introduce the subscript $c$, such that $x^{\ell}_{t,c}$ is the c'th token in the input sequence for learner $\ell$ at interaction $t$. The full training set is composed of all sequences for all interactions across all learners. Formally, we map each learner-timestep pair $(\ell,t)$ to a global index $i$, such that $x_i = x^{\ell}_{t}$ for some $\ell$ and $t$.

The resulting training dataset is denoted $\{ x_i \}_{i=1}^{N}$, where \( N \) is the total number of interactions across all students.

\paragraph{Fine-tuning}  

Fine-tuning large language models can be computationally expensive~\cite{xia2024understanding}. To address this, we employ Low-Rank Adaptation (LoRA)~\cite{hu2021lora}, which enables efficient adaptation by introducing a small set of trainable parameters \(\Theta\) while keeping the original model parameters \(\Phi_0\) frozen. Rather than updating the full parameter set, LoRA learns low-rank updates \(\Delta\Phi(\Theta)\) such that \(|\Theta| \ll |\Phi_0|\), significantly reducing memory and computation requirements while preserving model performance. The model is thus optimised to learn \(\Theta\) that maximise the log-likelihood of the training data  via \(\Phi_0 + \Delta\Phi(\Theta)\).

Let  $x_{i,c}$ refer to the token in position $c$ in the i'th training example and let $x_{i,<c} = (x_{i,1}\dots x_{i,c-1})$ denote all tokens prior to $x_{i,c}$. We fine-tune the model using a causal language modelling (CLM) objective in which the model autoregressively predicts $\hat{y}_{i,c}$, the token at position \(c\) in the \(i\)-th training sequence, conditioned on $x_{i,<c}$. The final loss is calculated between the models prediction $\hat{y}_{i,c}$ and the ground truth token $y_{i,c} = x_{i,c}$ . The optimisation objective is given in Equation~\ref{eq:lora_objective}.

\begin{equation} \label{eq:lora_objective}
  \max_{\Theta} \sum_{i=1}^{N} \sum_{c=1}^{|x_i|} m_{i,c} \log \left( P_{\Phi_0 + \Delta\Phi(\Theta)} \left( \hat{y}_{i,c} \mid x_{i,<c} \right) \right),
\end{equation}

where the masking variable \(m_{i,c}\) is defined as:

% \[
% m_{i,c} =
% \begin{cases}
% 1, & \text{\emph{if} NTKT-v where all tokens are predicted}, \\
% 1, & \text{\emph{if} NTKT-s where only tokens } \hat{a}^\ell_t = \hat{y}_{i,c}, \\
% 0, & \text{otherwise (ignored in loss)}.
% \end{cases}
% \]

\[
m_{i,c} =
\begin{cases}
% 1, & \text{\emph{when} fine-tuning NTKT-v models and } \forall \ y_{i,c}, \\
1, &  {y}_{i,c} \in a_t^\ell, \\
0, 
\end{cases}
\]

% \[
% m_{i,c} =
% \begin{cases}
% 1, & \text{NTKT-v: always}, \\[4pt]
% 1, & \text{NTKT-m: } x_{i,c}\in\{\text{Correct},\text{Incorrect}\}
%        \text{ and } x_{i,c}\text{ equals the tokenised form of } a^{\ell(i)}_{t(i)}, \\
% 0, & \text{otherwise (ignored)}.
% \end{cases}
% \]

% \paragraph{Vanilla NTKT-v (RQ1)}
% This variant follows the standard causal-language-modelling objective: every token
% $\hat{y}_{i,c}$ in the $i$\textsuperscript{th} training sequence $x_i$ is predicted
% autoregressively.  Using teacher forcing, the model sequentially generates the
% entire prompt–history block, the target question, its options, and the final
% answer-outcome tokens.  Consequently, the masking variable $m_{i,c}=1$ for all
% positions $c$, and the full sequence contributes to the loss.  This setup is used
% to answer \textbf{RQ1}.

\paragraph{NTKT with Selective Masking}
To isolate the predictive signal in the answer outcome alone, NTKT masks out
all tokens except those that spell \texttt{Correct} or \texttt{Incorrect}.
These outcome tokens are enclosed in \texttt{<cr>}\dots\texttt{</cr>} tags so
they can be located unambiguously in each sequence.  Formally, the loss is
computed only at positions $c$ for which
$x_{i,c}\in\{\text{Correct},\text{Incorrect}\}$; every other $c$ receives the
sentinel label $-100$ and is ignored.  Non-target tokens still participate in
the attention computation, providing full context, but exert no gradient
pressure.  This design is motivated by analogous masking techniques in speech
recognition~\cite{chen2024loss,jeon-etal-2025-prompt}.

\subsection{Evaluation Metrics}

The KT task is a binary classification problem. We use typical classification metrics to evaluate the predictive performance of the models. We use the F1-score, which captures the harmonic mean between the precision and the recall and the accuracy score as primary classification metrics. To measure the robustness of the probability values predicted from the models, we use the ROC-AUC metric.

We test the different models using a sequential experimental design, where the answer outcome $a^{\ell}_t$ is predicted using exercises $\mathcal{H}^\ell_t \in \{ e^\ell_1 \dots e^\ell_{t-1} \}$ and $e^\ell_t$ \cite{truelearn}. We use the hold-out validation approach for fine-tuning where parameters are learned on 90\% of the learners and the model is evaluated on the remaining learners.

\subsection{Experimental Design and Implementation}

We designed a sequence of three experiments to evaluate NTKT against established knowledge tracing baselines, following the structure of our research questions (RQs). Each experiment incrementally probes the model's predictive performance, data efficiency, and generalisation ability.

\paragraph{Model Variants and Architecture Selection.}
To assess the effect of model scale, we fine-tuned several open-source decoder-only LLMs from the LLaMA 3.2 family, specifically 1B and 3B parameter variants. These models serve as the foundation for NTKT, fine-tuned using the selective masking objective described earlier. The best-performing configuration from this stage is used in the subsequent ablation and cold-start analyses to ensure computational efficiency.

\paragraph{Experiment 1 – Benchmarking Predictive Performance (RQ1).}
The first experiment compares NTKT models with conventional deep learning baselines (DKT, AKT, AKT-text, and DTransformer) on the binary KT task. All models were trained and evaluated on the Eedi dataset using identical data splits, allowing for direct comparison. The goal is to evaluate whether fine-tuning pretrained LLMs with selective loss masking can surpass existing architectures in predictive accuracy and calibration.

\paragraph{Experiment 2 – Input Representation Ablation (RQ2).}
The second experiment examines how different levels of textual information influence NTKT performance. We modify the fine-tuning sequences under three input conditions:  
(1) \textbf{ID-only} (no text),  
(2) \textbf{ID + Concept tags}, and  
(3) \textbf{Full Question Text}.  
This design isolates the contribution of linguistic content to predictive performance, clarifying whether the performance gains stem primarily from the inclusion of question content.

\paragraph{Experiment 3 – Cold-Start Generalisation (RQ3).}
The final experiment investigates NTKT’s capacity to generalise under two cold-start scenarios:
\begin{itemize}[noitemsep, leftmargin=*]
    \item \textbf{User Cold Start:} Models are evaluated on learners unseen during training. We compute the average F1 score at each timestep across these held-out users, tracking how quickly each model personalises to new learners as their interaction history grows.
    \item \textbf{Question Cold Start:} Models are evaluated on questions withheld entirely during training. Ten such unseen items were selected to span a representative range of difficulty and topic diversity, with 7,297 students (18.5\% of the total) contributing responses. This setup measures robustness to novel instructional content.
\end{itemize}

\paragraph{Implementation Details.}
All LLM fine-tuning was conducted with identical hyperparameter settings to ensure comparability. Training used an NVIDIA A100 80\,GB GPU with five random initialisations. Models were fine-tuned for up to 20{,}000 steps with early stopping triggered when the evaluation loss failed to improve by at least 0.001 over ten evaluation intervals (evaluated every 250 steps). Configuration details include: maximum sequence length = 15{,}000 tokens, LoRA rank = 16, LoRA $\alpha$ = 16,  per-device batch size = 4, gradient accumulation = 4, learning rate = 2$\times$10$^{-4}$ with cosine scheduling and 50 warmup steps, and AdamW (8-bit) optimisation with weight decay = 0.01. Cross-entropy loss was used throughout.

To ensure reproducibility, we used fixed random seeds, a consistent tokenisation pipeline, and released all preprocessing and evaluation code in the supplementary material.

\section{Results and Discussion}

In this section, we present the outcomes of our experiments and discuss their implications in relation to the research questions.

\subsection{Predictive Performance Comparison (RQ1)}

Table~\ref{tab:performance_eedi} summarises the overall predictive performance across all models. NTKT consistently achieves the highest results on the binary Knowledge Tracing task, outperforming all traditional baselines across F1, Accuracy, and AUC metrics.

Among the evaluated configurations, the LLaMA 3B NTKT model achieves the best overall performance (F1 = 85.22\%, Accuracy = 78.87\%, AUC = 90.32\%), followed by the 1B variant. These results demonstrate that pretrained language model representations can be effectively adapted to model student learning trajectories, even at relatively small model scales.  

Compared with strong neural baselines such as DKT, AKT, and DTransformer, NTKT shows substantial gains—exceeding 10 percentage points in AUC over the best-performing baseline. This indicates that incorporating question text and leveraging pretrained representations provides richer contextual understanding of student–question interactions, enabling more accurate prediction of student performance.  

Notably, even the smallest NTKT model (1B) surpasses all baselines, suggesting that the architecture generalises well and is efficient in capturing relevant behavioural and linguistic patterns without requiring massive model capacity.
\begin{table}[ht]
\centering
\caption{Predictive Performance on the Eedi Dataset. Results show Mean $\pm$ SD across 5 runs. Best results are \textbf{bolded}.}
\label{tab:performance_eedi}
\begin{tabular}{lccc}
\toprule
\textbf{Model} & \textbf{F1} & \textbf{Accuracy} & \textbf{AUC} \\
\midrule
\textit{Baselines} & & & \\
DKT & 77.46 $\pm$ .0000 & 63.22 $\pm$ .0000 & 73.23 $\pm$ .0013 \\
AKT & 76.92 $\pm$ .0043 & 68.31 $\pm$ .0010 & 72.37 $\pm$ .0013 \\
AKT\textsubscript{text} & 77.02 $\pm$ .0008 & 68.90 $\pm$ .0008 & 72.23 $\pm$ .0031 \\
DTransformer & 76.80 $\pm$ .0020 & 68.88 $\pm$ .0012 & 73.23 $\pm$ .0076 \\
No-FT & 74.25 $\pm$ .0030 & 66.78 $\pm$ .0020 & 75.60 $\pm$ .0030 \\
\midrule
\textit{NTKT (Ours)} & & & \\
LLaMA 1B & 82.11 $\pm$ .0023 & 77.23 $\pm$ .0010 & 89.10 $\pm$ .0010 \\
LLaMA 3B & \textbf{85.22 $\pm$ .0013} & \textbf{78.87 $\pm$ .0020} & \textbf{90.32 $\pm$ .0014} \\
\bottomrule
\end{tabular}
\end{table}

\subsection{Impact of Input Representation (RQ2)}
\label{sec:inputrep}

Table~\ref{tab:ntkt_llama3b_oriented} shows how varying levels of textual information influence NTKT performance. In all cases, including the full question text yields the best results, confirming that natural language content provides essential pedagogical cues that enhance predictive accuracy and calibration.

The concept-only configuration, which retains semantic tags but omits full text, leads to a decline across all metrics (e.g., AUC dropping from 90.32\% to 84.59\%). Meanwhile, the ID-only condition, where models receive only symbolic question identifiers, retains competitive F1 but exhibits significant AUC degradation (77.06\%), suggesting less reliable probability estimates when textual context is removed.  

These findings confirm that textual information substantially strengthens NTKT’s ability to capture question semantics and their relationship with student responses, validating the importance of incorporating linguistic context into knowledge tracing.

\begin{table}[t]
\centering
\footnotesize
\caption{Eedi NTKT (LLaMA 3B) performance using different feature representations. Best and second-best results are shown in \textbf{bold} and \emph{italic}, respectively.}
\label{tab:ntkt_llama3b_oriented}
\begin{tabular}{l|ccc}
\toprule
\textbf{Features} & \textbf{F1} & \textbf{Accuracy} & \textbf{AUC} \\
\midrule

ID-only &
79.16 ± .0131 &
69.22 ± .0120 &
77.06 ± .0085 \\

Concept-only &
\emph{83.22 ± .0114} &
\emph{76.77 ± .0105} &
\emph{84.59 ± .0062} \\

Full Text &
\textbf{85.22 ± .0013} &
\textbf{78.87 ± .0020} &
\textbf{90.32 ± .0014} \\

\bottomrule
\end{tabular}
\end{table}

\subsection{Generalisation to Cold-Start Scenarios (RQ3)}
\label{sec:generalisation}

\begin{figure}[t]
\centering
\includegraphics[width=.9\linewidth]{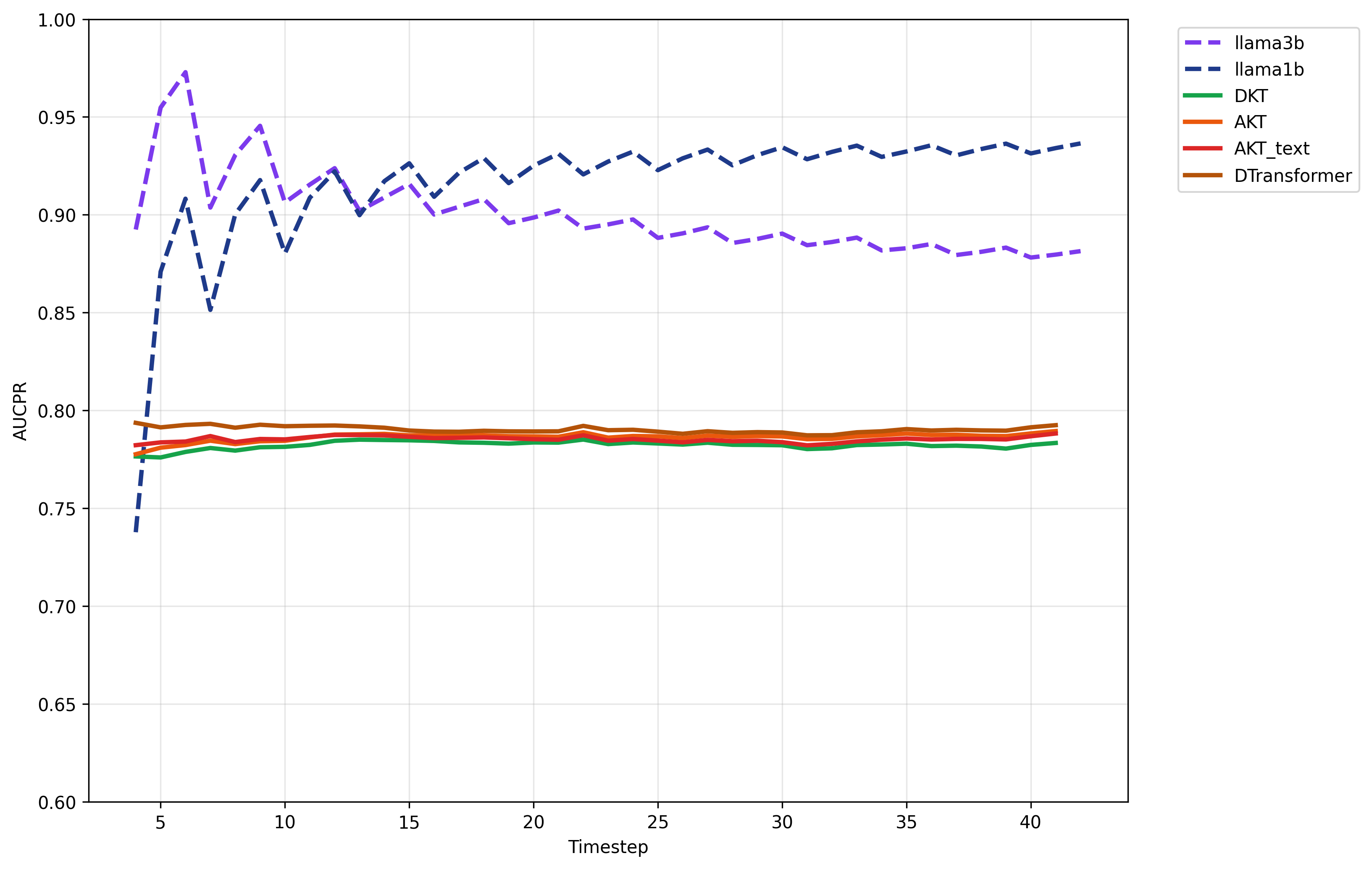}
\caption{Average F1 score across timesteps for NTKT and baseline models in user cold-start scenarios. Curves show average F1 across all held-out users, from the initial interaction through subsequent steps.}
\label{fig:coldstart}
\end{figure}

\paragraph{User Cold Start.}
Figure~\ref{fig:coldstart} presents results for user cold-start conditions. NTKT models achieve significantly higher F1 scores than baselines from the very first interaction and continue improving rapidly as additional responses become available. This indicates that NTKT captures generalisable representations of question semantics and student behaviour, enabling fast personalisation to new learners. Baseline models, by contrast, show slower and more gradual improvement, reflecting their dependence on accumulating user-specific data.

\begin{figure}[t]
\centering
\includegraphics[width=.8\linewidth]{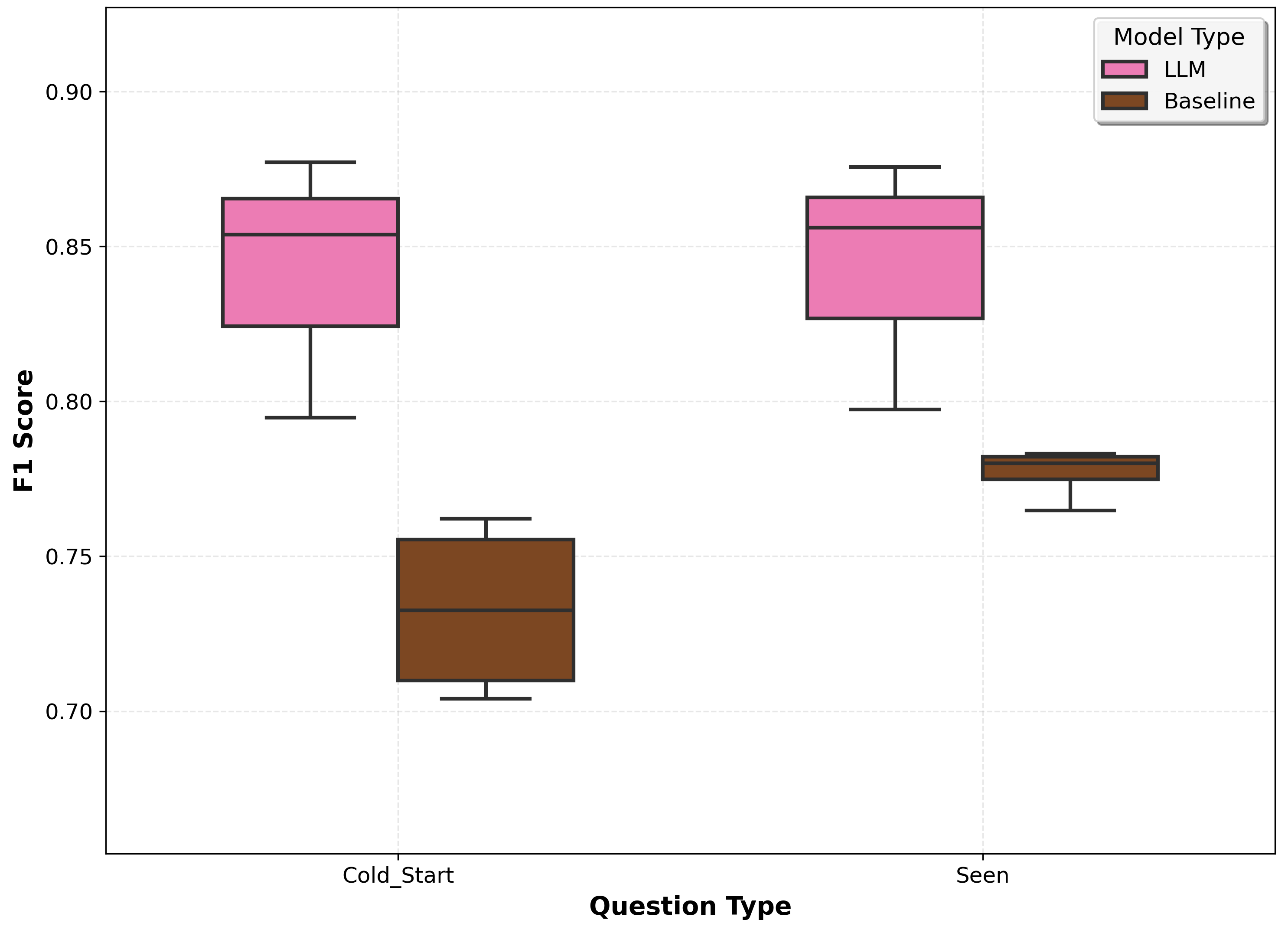}
\caption{Performance of NTKT and baseline models on seen versus cold-start questions, showing stable generalisation across unseen content.}
\label{fig:coldstart_q}
\end{figure}

\paragraph{Question Cold Start.}
Cold-start robustness for unseen questions is shown in Figure~\ref{fig:coldstart_q}. Baseline models exhibit a clear performance drop between seen and unseen questions (0.777 to 0.732, $p<0.001$), whereas NTKT maintains stable performance (0.843 on both, $p>0.5$). Both NTKT model variants show stable or slightly improved gains on unseen items. This demonstrates strong generalisation to new content, driven by NTKT’s ability to leverage pretrained representations and linguistic cues rather than memorised identifiers.

\section{Conclusion, Limitations and Future Work}

This paper introduces Next Token Knowledge Tracing (NTKT), a novel LLM-based approach that reformulates knowledge tracing as a next token prediction task. NTKT is a framework that was developed to overcome the limitations posed by existing Deep Knowledge Tracing models and their exclusion of question text. The NTKT framework contributes towards the AI education community by providing better performing KT models, that can generalise well in cold start settings. Furthermore, we highlight the value of question text for student modelling research and an approach to effectively make use it.

These contributions are supported through three core experiments. First, we showcased the performance of NTKT models against existing baselines on a widely used KT dataset. Next, To ensure this higher performance was due to question text, our ablation studies isolated the importance of complete textual information in knowledge tracing. Finally we presented results from our cold-start experiment. This experiment demonstrated NTKT's generalization capabilities across student and question cold-start scenarios. 

\subsection{Limitations}

The present study has four main limitations.
(i) Computational cost – NTKT relies on substantially larger foundation models than traditional KT, so both training and inference require more GPU memory, time and energy, even when parameter-efficient fine-tuning is used.
(ii) Eedi – We used the Eedi dataset due to its size, provision of question text and recognition amongst the AI Education community as a KT benchmark. However, we recognise that validating NTKT across numerous datasets is necessary for widespread adoption. Also, Eedi is restricted to the mathematics domain in the English language. Our experiments provide little hard evidence on NTKT's value to multi-domain, multi-lingual scenarios.
(iii) Quantization – all experiments were performed with 4-bit weights to fit hardware constraints; this compression is known to introduce some loss of numerical precision and may reduce accuracy.
(iv) Data requirements – NTKT assumes rich, high-quality question text. In domains where such content is unavailable, proprietary, or heavily noise-corrupted, the advantages we observe are likely to diminish.

\subsection{Future Work}

Several promising directions exist for future research. These include investigating cross-domain adaptation to other subject areas and multi-lingual settings. The framework could also be expanded to handle multi-modal content like mathematical equations and diagrams for STEM education. Another key direction is developing interpretability tools to extract actionable pedagogical insights from attention patterns as well as investigating sources of potential bias within NTKT so that the outputs can be presented to learners via intelligent user interfaces for learning (e.g. like X5Learn \cite{perez2022watch}).  Furthermore, NTKT's generative capabilities could be leveraged to provide personalized questions \cite{li2025novel} and explanations based on predicted behavior. Future work should also explore computational optimizations to make NTKT more accessible for resource-constrained educational environments while maintaining its performance advantages.
\bibliography{aaai2026}

\clearpage
\appendix
\section{Prompt Templates and Implementation Details}
\label{sec:appendix_prompts}

This appendix provides the full textual structure used to prompt the Large Language Model for the Next Token Knowledge Tracing (NTKT) task.

\begin{listing*}[t]
\begin{lstlisting}[caption={Instruction prompt template ($\mathcal{P}_{\texttt{LLM}}$ in Figure~\ref{fig:pipeline}) used for NTKT fine-tuning.}, label={lst:ntkt_prompt}]
Given the following student question and answer history, predict whether the student will answer the target question correctly or incorrectly. The target question is enclosed in <target> tags and the options are enclosed in <options> tags. Respond with ``Correct'' if you think they will answer correctly, or ``Incorrect'' if you think they will answer incorrectly.

<history>:
<Q>
  <text>[Question 1]</text>
  <options>[Options A-D]</options>
  <QID>[ID]</QID>
  <C>[Concept]</C>
</Q><cr>[Correct/Incorrect]</cr> 
...
<Q>
  <text>[Question N]</text>
  <options>[Options A-D]</options>
  <QID>[ID]</QID>
  <C>[Concept]</C>
</Q><cr>[Correct/Incorrect]</cr>
</history>

What do you predict they will answer for the target question: 
<target>
  <text>[Target Question]</text>
  <options>[Option A-D]</options>
  <QID>[ID]</QID>
  <C>[Concept]</C>
</target>:
\end{lstlisting}
\end{listing*}

\end{document}